\documentclass[lettersize,journal]{IEEEtran}
\usepackage{amsmath,amsfonts}
\usepackage{algorithmic}
\usepackage{algorithm}
\usepackage{array}
\usepackage[caption=false,font=normalsize,labelfont=sf,textfont=sf]{subfig}
\usepackage{textcomp}
\usepackage{stfloats}
\usepackage{url}
\usepackage{verbatim}
\usepackage{graphicx}
\usepackage{cite}
\usepackage{threeparttable}
\usepackage{enumitem}
\usepackage{booktabs}
\usepackage{amsmath}
\usepackage{extarrows}
\usepackage{multirow}
\usepackage{makecell}
\usepackage{xcolor}

\begin{document}

\title{Where to Go Next Day: Multi-scale Spatial-Temporal Decoupled Model for Mid-term Human Mobility Prediction}

\author{Zongyuan Huang{$^\dag$},~Weipeng Wang{$^\dag$},~Shaoyu Huang,~Marta C. Gonz\'alez,~Yaohui Jin{$^\ast$},~\IEEEmembership{Member,~IEEE},~Yanyan Xu{$^\ast$}
\IEEEcompsocitemizethanks{
\IEEEcompsocthanksitem Z. Huang, W. Wang, S. Huang, Y. Jin, and Y. Xu are with 
(1) the MoE Key Laboratory of Artificial Intelligence and AI Institute, Shanghai Jiao Tong University, Shanghai 200240, China, (2) the Data-Driven Management Decision Making Lab, Shanghai Jiao Tong University, Shanghai 200240, China, 
and (3) the School of Computer Science and Engineering, Shanghai Jiao Tong University, Shanghai 200240, China.. E-mail: \{herozen, weipeng001, leak\_ish, jinyh, yanyanxu\}@sjtu.edu.cn. 
\IEEEcompsocthanksitem M. C. González is with (1) the Department of Civil and Environmental Engineering, University of California, Berkeley, CA 94720, USA, (2) the Laboratory for DCRP, University of California, Berkeley, CA 94720, USA, 
and (3) the Energy Technologies Area, Lawrence Berkeley National Laboratory, Berkeley, CA 94720, USA. E-mail: martag@berkeley.edu. \\
$^\dag$ Equal contribution. \\
$^\ast$ Corresponding authors.
}
}


\markboth{Journal of \LaTeX\ Class Files,~Vol.~14, No.~8, August~2021}%
{Shell \MakeLowercase{\textit{et al.}}: A Sample Article Using IEEEtran.cls for IEEE Journals}


\maketitle

\begin{abstract}

Predicting individual mobility patterns is crucial across various applications. While current methods mainly focus on predicting the next location for personalized services like recommendations, they often fall short in supporting broader applications such as traffic management and epidemic control, which require longer period forecasts of human mobility. This study addresses mid-term mobility prediction, aiming to capture daily travel patterns and forecast trajectories for the upcoming day or week. We propose a novel Multi-scale Spatial-Temporal Decoupled Predictor (MSTDP) designed to efficiently extract spatial and temporal information by decoupling daily trajectories into distinct location-duration chains. Our approach employs a hierarchical encoder to model multi-scale temporal patterns, including daily recurrence and weekly periodicity, and utilizes a transformer-based decoder to globally attend to predicted information in the location or duration chain. Additionally, we introduce a spatial heterogeneous graph learner to capture multi-scale spatial relationships, enhancing semantic-rich representations. Extensive experiments, including statistical physics analysis, are conducted on large-scale mobile phone records in five cities (Boston, Los Angeles, SF Bay Area, Shanghai, and Tokyo), to demonstrate MSTDP's advantages. Applied to epidemic modeling in Boston, MSTDP significantly outperforms the best-performing baseline, achieving a remarkable 62.8\% reduction in MAE for cumulative new cases. 

\end{abstract}


\section{Introduction}
~\label{sec:intro}

Human mobility modeling has attracted considerable attention in fields such as commercial planning, healthcare, urban management, and mobile and network applications~\cite{wu2018location,xu2018planning,zhou2023predicting,jia2023human}. With the rapid advancements in technology, diverse mobility data are now being collected from multiple sources, driving innovation and application across these domains. For example, location-based social network (LBSN) data from check-in services like Yelp and Foursquare enable targeted, location-aware advertising~\cite{xu2020survey}. Mobility trajectories obtained from mobile devices are instrumental in analyzing and controlling the spread of epidemics~\cite{jia2020population,luca2022modeling,xu2023urban}, while GPS data from vehicles plays a crucial role in optimizing traffic flow and management~\cite{chen2014b,xu2021understanding,ji2022spatio}. These developments underscore the growing importance of accurate mobility models in addressing complex real-world challenges.

Human mobility prediction is generally addressed at two scales: collective and individual~\cite{luca2021survey}. This study concentrates on the individual scale, with a specific emphasis on predicting personal mobility patterns. The core challenge in individual mobility prediction revolves around next-location forecasting, which aims to predict an individual's future destinations based on their historical movement data. Mobility trajectories are commonly represented as sequences of locations, and the primary task is to extract meaningful transition patterns from these sequences. By identifying regularities and trends in an individual’s movement, this approach seeks to enhance the precision of mobility predictions. Traditional models rely on Markov chains~\cite{feng2015personalized}, while deep learning models apply advanced techniques such as Recurrent Neural Networks (RNNs) and Transformers~\cite{guo2020attentional,zhang2022next,yang2022getnext}. 
Owing to the spatial characteristics of locations, researchers have explored specific modules to capture spatial relationships. 
Recent advancements in Graph Neural Networks (GNNs)~\cite{wang2022learning,lim2022hierarchical,qin2023disenpoi,yan2023spatio,jin2023spatio,zhou2025disentangled,song2025integrating} and Large Language Models (LLMs)~\cite{zhong2025comapoi,chen2025NextPOI} have significantly enhanced human mobility prediction by capturing spatial-temporal dynamics and leveraging contextual sequential information. Their integration offers promising opportunities for more accurate and comprehensive modeling of mobility behaviors.

\begin{figure}
    \centering
    \includegraphics[width=0.45\textwidth]{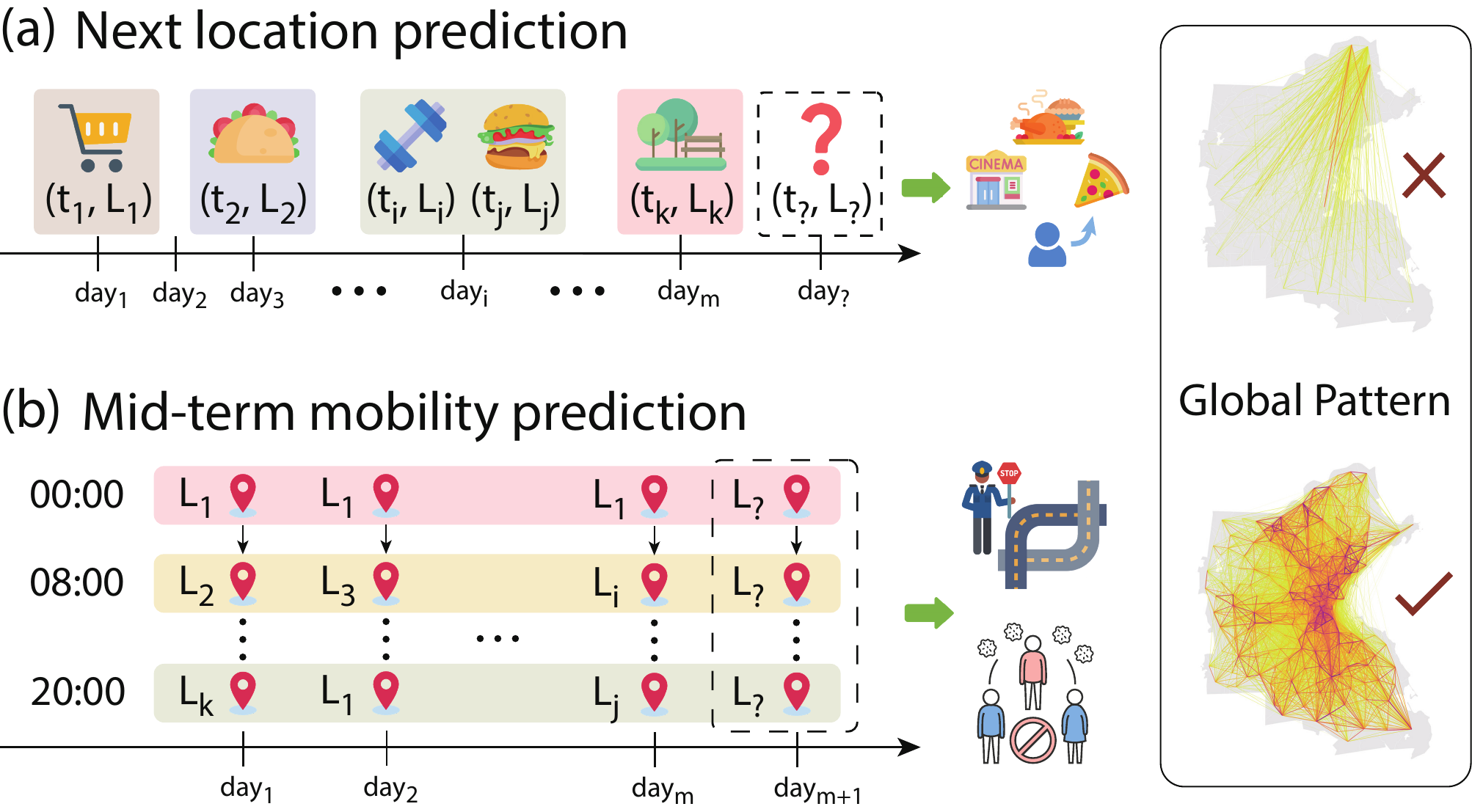}
    \caption{Task illustrations of next location prediction and mid-term mobility prediction.}
    \label{fig:motivation}
\end{figure}

Predicting the next location is a widely studied task in human mobility analysis, primarily applied in personalized services such as point-of-interest (POI) recommendation and navigation assistance, as illustrated in Figure~\ref{fig:motivation}(a). These tasks focus on identifying short-term, localized movement patterns, such as determining an individual's immediate next stop. Given their focus, next location prediction tasks typically rely on datasets with sparse and irregular temporal resolutions, which reflect the episodic nature of the movements being modeled.
In contrast, urban management applications, such as travel demand forecasting and epidemic transmission modeling, require predicting comprehensive daily trajectories for residents. This task, referred to as mid-term mobility prediction, emphasizes modeling complete daily travel behaviors, as shown in Figure~\ref{fig:motivation}(b). Unlike next location prediction, mid-term mobility prediction is driven by the need to address broader societal challenges, such as optimizing urban planning, managing traffic flows, and assessing public health risks. To meet these objectives, it is essential to capture an individual's full-day mobility patterns, which necessitates dense observation data capable of capturing sequential and structural continuity throughout the day.
While typical next location prediction models have been adapted to mid-term mobility prediction tasks, they often rely on iterative mechanisms that predict one trip at a time by incrementally combining predicted outcomes with historical data to forecast subsequent trips. Existing approaches encounter significant obstacles in mid-term mobility prediction tasks, primarily due to their inability to effectively handle repeated consecutive locations and overly long sequences. Such issues not only challenge pattern recognition and memory capabilities but also result in compounding prediction errors over extended time horizons. Furthermore, these methods fail to incorporate multi-scale spatiotemporal patterns that are essential for capturing individuals' daily travel behaviors in a holistic and realistic manner. Motivated by these limitations, we propose a novel framework specifically designed for the mid-term mobility prediction task, offering a robust and comprehensive solution to address the unique challenges posed by this application.

\begin{figure}
    \centering
    \includegraphics[width=0.42\textwidth]{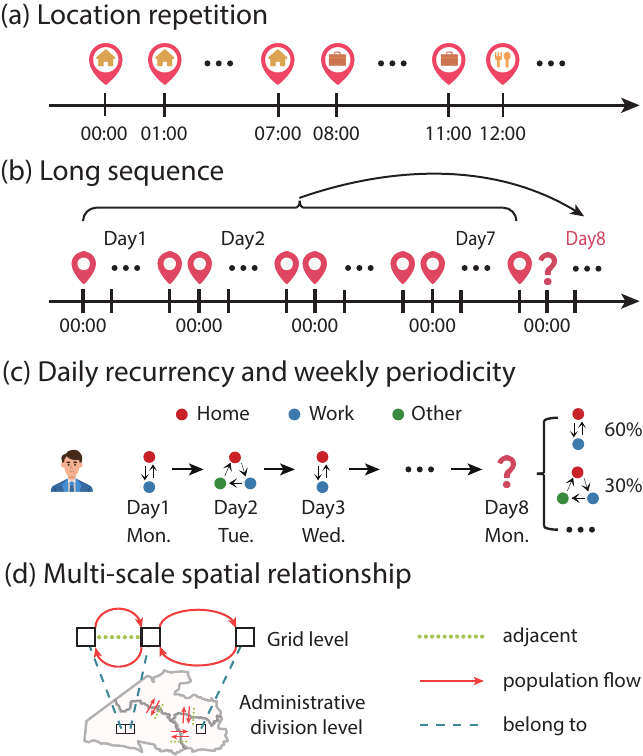}
    \caption{Characteristics of mid-term mobility prediction.}
    \label{fig:challenges}
\end{figure}

The characteristics of mid-term mobility prediction are illustrated in Figure~\ref{fig:challenges}. 
Firstly, accurate prediction of an individual's daily trajectory requires knowing their location at each time interval. However, a prolonged stay at a specific location results in consecutive repetitions of the same location within the location sequence. For instance, hourly observations often show repeated instances of home occupancy during the nighttime and early morning, as illustrated in Figure~\ref{fig:challenges}(a). Such consecutive location repetitions in trajectory sequences confound the sequence models in accurately identifying location transitions. 
Secondly, learning the daily travel patterns over multiple days is essential for accurately predicting future behaviors. Yet, the long location sequences pose challenges to the perception and memory capabilities of sequence models, as indicated in Figure~\ref{fig:challenges}(b). 
Thirdly, people's daily travel activities exhibit multi-scale regular patterns, encompassing both daily recurrency and weekly periodicity. Daily recurrency refers to the consistency and similarity observed in individual travel behaviors across consecutive days. Urban residents' daily travel often concentrates on a limited set of primary patterns, such as commuting between regular locations. For illustrative purposes, Figure~\ref{fig:challenges}(c) highlights a conceptual example, showing that a significant proportion of workday trips involve commuting directly between home and work, while a smaller proportion includes additional stops before returning home. It should be noted that the percentages (e.g., 60\% and 30\%) are not based on empirical data but are presented solely as illustrative examples to emphasize the regularity that underpins mobility patterns. On the other hand, weekly periodicity captures cyclical patterns in travel behaviors, such as the recurrence of similar journeys on Mondays. These regular patterns provide essential insights for the modeling of daily trajectories.
Lastly, cities display diverse spatial relationships among their regions, encompassing proximity, population flows, and significant administrative connections, as exemplified in Figure~\ref{fig:challenges}(d). Multiscale spatial granularity provides diverse insights into urban structural features, aligning with the multiscale characteristics encompassing both short and long-distance travel. Investigating these multiscale spatial relationships can enhance location representations and improve model performance in downstream tasks.

In this paper, we propose a Multi-Scale Spatial-Temporal Decoupled Predictor (MSTDP) for the mid-term mobility prediction task. Our approach starts by decoupling individuals' daily trips into separate location and duration chains, thereby eliminating consecutive repetitions and directly capturing the daily transition patterns. Following this, a two-layer hierarchical encoder is employed to extract multi-scale temporal patterns like daily recurrency and weekly periodicity, efficiently broadening the receptive field and capturing longer-term behavioral regularities. Subsequently, a Transformer-based decoder integrates encoded historical information and previously predicted partial results to forecast trajectories for the following day. Furthermore, we introduce a heterogeneous graph module to capture the spatial and accessibility relationships between regions at different scales, thereby enriching the semantic representation of locations and enhancing predictive accuracy.
We evaluate MSTDP against nine baselines using five urban-scale mobile record datasets. Empirical results demonstrate the superior performance of the proposed model. We further conduct comprehensive statistical physics analyses using Boston as a case study, examining motifs, travel distances, origin-destination flows, and other dimensions to provide a multidimensional assessment of the model's effectiveness.
Moreover, we utilize the model's predictions to analyze the epidemic transmission dynamics. Specifically, we simulate the epidemic spread using the SEIR model~\cite{chang2021mobility} and the population transfer probability matrices derived from travel trajectories. This enables monitoring of the cumulative new cases and the current active cases. Compared to the baseline, MSTDP demonstrated remarkable improvements with reductions in MAE by 87.6\% for the cumulative new cases and 62.8\% for the current active cases, highlighting the superior predictive capability of our model in modeling travel behaviors.

We summarize the key contributions as follows:
\begin{itemize}[leftmargin=*]
    \item We highlight mid-term mobility prediction, a crucial task distinct from next-location prediction. 
    This task concentrates on the daily mobility patterns of urban populations, supporting applications in urban management such as disease control and traffic regulation.
    \item We propose a novel Multi-scale Spatial-Temporal Decoupled Predictor (MSTDP), employing a spatial-temporal decoupling approach to address repeated observations and hierarchical encoders to efficiently capture the multi-scale temporal patterns, including daily recurrence and weekly periodicity of travel behavior.
    \item We introduce a novel spatial heterogeneous graph to describe the multi-scale location relationships, including mobility flow and adjacent relationships across regions at different scales. To capture these relationships, we develop a graph network-based heterogeneous geospatial module, enhancing the semantic richness of location representations. 
    \item Extensive experiments on five urban datasets validate the effectiveness of the proposed model. We provide diverse statistical physics analysis perspectives for comprehensive evaluation. Moreover, we apply the model's predictions to analyze the epidemic transmission in Boston, demonstrating the significant advantages of MSTDP.
\end{itemize}

\section{Related Work}
\subsection{Next location prediction}
Existing work on individual mobility prediction primarily focuses on next location prediction, which typically utilizes sequential models to capture transition relationships between locations. Early methods mainly rely on Markov chains to construct location transition probability matrices, with techniques to incorporate temporal and spatial information to capture individual travel preferences~\cite{cheng2013you,feng2015personalized,jiang2016timegeo}.
The evolution of deep learning has effectively facilitated the modeling of complex interactions.
The foundational framework of location sequence models has progressed from the RNN series~\cite{feng2018deepmove,guo2020attentional,xu2022metaptp} to attention-based Transformer architectures~\cite{yang2022getnext,zhang2022next}, consequently enhancing the capacity to capture high-order transition patterns.
In addition to the transition regularity, location sequences offer a wealth of spatiotemporal information. 
Thus, researchers have introduced innovative mechanisms from diverse perspectives, such as temporal and spatial semantics, to enhance sequence models~\cite{sun2020go,luo2021stan,lian2020geography,lin2021pre,wang2022modeling,qin2022disentangling,huang2024learning,jiang2024towards,xu2024taming}. Besides, some work also incorporates location category information to further constrain the location selection~\cite{yu2020category,dong2021exploiting,huang2024human}. 

Recent advancements have also explored leveraging Large Language Models (LLMs) to capture rich contextual semantics~\cite{zhong2025comapoi,chen2025NextPOI}, adopting Test-Time Adaptation techniques to dynamically adjust the model based on test data~\cite{han2025adamove}, and utilizing hyperbolic space to model hierarchical relationships~\cite{qiao2025Hyperbolic}. Moreover, some studies have incorporated crowd flow information to integrate group-level travel dynamics, further complementing individual mobility modeling~\cite{bontorin2025mixing,long2025universal}. 

However, methods designed for next location prediction face several challenges when applied to the mid-term mobility prediction task. As discussed in Introduction~\ref{sec:intro}, the presence of repeated consecutive locations and excessively long sequences poses significant challenges to their pattern recognition and memory capabilities. Furthermore, these methods fail to effectively capture the multi-scale spatiotemporal patterns inherent in individuals’ daily travel behaviors.
To mitigate their limitations, we propose a spatiotemporal decoupled hierarchical framework, including a heterogeneous graph learner, which enlarges the perception field and efficiently captures the multi-scale spatiotemporal information.

Over the past few years, graph neural networks have evolved rapidly and have been widely applied to study human mobility~\cite{han2020stgcn,lim2022hierarchical,yin2023next,wang2023adaptive,wang2023eedn,xu2023revisiting}. Some approaches leverage user-POI heterogeneous graphs to model the interaction between individuals and locations, effectively capturing personalized mobility patterns and spatial dynamics~\cite{qiao2025Hyperbolic,song2025integrating}. In contrast, other studies construct flow graphs and adjacency graphs to derive distinct representations of locations based on spatial-temporal relationships, devising mechanisms to combine them for enhanced prediction performance~\cite{wang2022learning,wang2022graph,qin2023disenpoi,zhou2025disentangled}.
In this work, we take a step further by introducing a heterogeneous graph that integrates flow, adjacency, and inclusion relationships to learn a more comprehensive and informative location representation. Unlike existing approaches, our method incorporates multi-level adjacency and flow relationships, enabling the efficient modeling of complex geographic spatial structures to better capture human mobility dynamics.

\subsection{Successive location prediction}
Successive location prediction focuses on forecasting sequences of future visited locations based on historical trajectories combined with the known timing of location transitions~\cite{luo2022rlmob,wang2023multi,kim2023cell}. Future transition times provide travel schedule information that simplifies the prediction task. However, such information is typically unavailable in real-world scenarios. In contrast, mid-term mobility prediction aims to infer future mobility patterns solely based on historical behaviors, without relying on predefined transition times. This task is application-driven, particularly in urban management contexts, and emphasizes time structures that align with habitual travel patterns, such as daily or weekly intervals. For example, predicting population travel behaviors for the next day is crucial for applications like traffic planning or epidemic control.
Although some prior studies also focus on forecasting ``next-day" mobility, they often broadly categorize these tasks under location prediction without explicitly defining or distinguishing mid-term mobility prediction as a unique problem~\cite{li2020hierarchical}. In this work, we aim to provide a clear definition of mid-term mobility prediction, differentiate it from successive location prediction, and highlight its significance in supporting real-world applications.

\section{Preliminary}
\label{sec:pre}
In this section, we first formulate the next day mobility prediction task, and then introduce the definition of two basic location graphs.

\subsection{Individual Trajectory} 

Analyzing individual mobility patterns throughout the day necessitates dense data. To this end, we leverage Call Detail Record (CDR) data, which records locations using latitude and longitude coordinates rather than Points of Interest (POI) as in LBSN data. To address privacy concerns and simplify the modeling process, researchers partition the urban area into grids of equal size and map individuals' trajectories to grids. Each grid is treated as a discrete location visited by individuals~\cite{jiang2016timegeo}. The grid size used in this study is 1 km $\times$ 1 km. Here, the location set is denoted as $\mathcal{L} = \{l_1, \dots, l_{|\mathcal{L}|}\}$. We then partition each day into fixed-size timeslots. Using an hourly time window as an example, the daily trajectory is expressed as ${\rm D} = \{l^0_p, \dots, l^{23}_q\}$, where $l^t_p$ denotes the stay location $l_p$ within the hour $[t,t+1), t \in \mathbb{N}$. Let $\mathcal{U} = \{u_1, \dots, u_{|\mathcal{U}|}\}$ denotes a set of individuals. Then all of trajectories of an individual $u$ can be represented as ${\rm TRAJ}_u = \{{\rm D}_1, \dots, {\rm D}_{\rm N}\}$, where ${\rm N}$ is the number of total trajectories of the user, ${\rm N}=|{\rm TRAJ}_u|$. 

\textbf{Task Formulation.} Given an individual $u$ and his or her historical trajectories $\{{\rm D}_1, \dots, {\rm D}_{k} \} $, the goal is to predict the next day trajectory $\hat{\rm D}_{k+1} = \{ \hat{l}^0_p, \dots, \hat{l}^{w}_q \}$, where $k \leq |{\rm TRAJ}_u|$ -1. $w$ is determined by the size of the time window. In the methodology section, we present the model's details using hour-long windows as an example. In the experiments, different datasets require varying time window sizes, selected based on the level of data sparsity.

\subsection{Basic Location Graph}

The flow graph and adjacency graph have been used in recent studies to capture the spatial relationships of locations~\cite{lim2022hierarchical,wang2022learning,qin2023disenpoi}. We define these two graphs as follows.

\textbf{Flow Graph} is a directed graph $G^f = (V, E^f)$ to describe transition relationships among locations, where the node $v_j, v_j \in V$, denotes a location, the edge $e^f_{pq}, e^f_{pq} \in {E^f}$, represents the number of trips from $v_p$ to $v_q$ produced by all individuals. In this study, we only count $E^f$ in the training set.

\textbf{Adjacency Graph} is an undirected graph $G^a = (V, E^a)$ to depict the proximity between locations. The node set $V$ aligns with the location set in the flow graph $G^f$. The undirected edge $e^a_{pq}$ in the edge set $E^a$ indicates that the geospatial distance between $v_p$ and $v_q$, denoted as $dist_{pq}$, is less than a pre-defined threshold $\bar{d}$. In our work, $\bar{d}$ is set to 2~km for a grid with 1~km width. $e^a_{pq}=1$ if $dist_{pq} \leq \bar{d}$ and $e^a_{pq}=0$ otherwise.

\begin{figure*}
    \centering
    \includegraphics[width=0.95\textwidth]{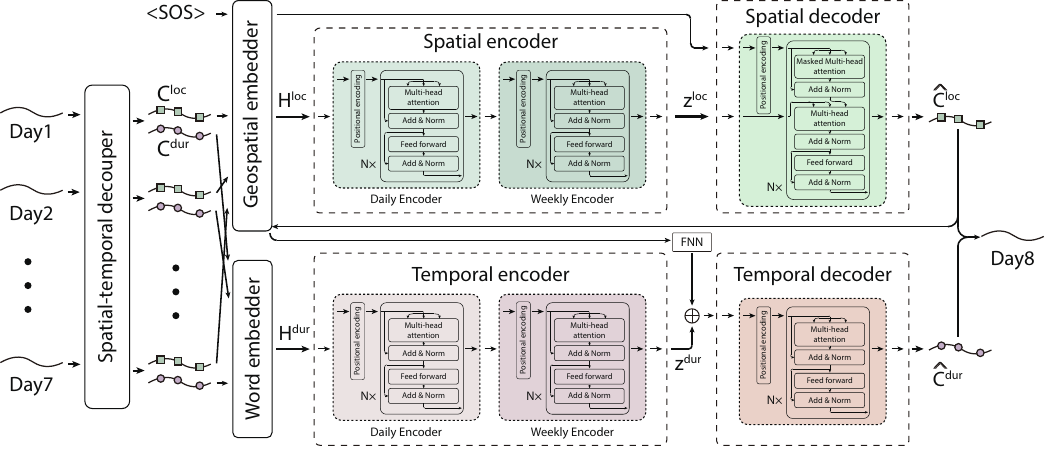}
    \caption{The framework of the proposed Multi-scale Spatial-Temporal Decoupled Predictor (MSTDP).}
    \label{fig:framework}
\end{figure*}

\section{Methodology}

In this section, we will sequentially introduce the four primary components of the Multi-scale Spatial-Temporal Decoupled Predictor (MSTDP), as illustrated in Figure~\ref{fig:framework}. 

\subsection{Spatial-Temporal Decoupler}

Modeling the daily travel behavior of individuals at the urban scale presents challenges of location repetition and excessively long sequences, as mentioned in Introduction~\ref{sec:intro}. Location repetition arises from individuals being observed multiple times at the same location due to their prolonged stays. Therefore, we can extract the stay duration from a daily trajectory to obtain the location sequence that eliminates consecutive repetitions. Specifically, the proposed spatiotemporal decoupler separates the trajectory of the $k$-th day ${\rm D}_k$ into a location chain ${\rm C}^{\rm loc}_k=\{l_p, \dots, l_q\}$ and a duration chain ${\rm C}^{\rm dur}_k=\{t_1, \dots, t_n\}$, where $t \in \mathbb{N}^+$ and $\sum_{s=1}^{n}t_s=24$. The elements $t_i$ in the duration chain correspond to the stay duration in the $i$-th location of ${\rm C}^{\rm loc}_k$. Both chains are of equal lengths $|{\rm C}^{\rm loc}_k| = |{\rm C}^{\rm dur}_k|$. Taking a daily trajectory $\{l^0_1, l^1_1, l^2_1, l^3_2, l^4_2, \dots, l^{21}_3, l^{22}_1, l^{23}_1\}$ as an example, the decoupled location and duration chains would be $\{l_1,l_2,\dots,l_3,l_1\}$ and $\{3,2,\dots,1,2\}$ respectively. In this way, we can disentangle lengthy sequences with consecutively repeated locations into two shorter chains, effectively reducing the complexity of pattern learning. 
Moreover, the decoupled location chain can be viewed as a location-specific motif~\cite{schneider2013unravelling} that enables the model to learn the integrity of an individual's daily travel patterns and the cyclical nature of their everyday behavior.

\subsection{Geospatial Embedder}
\label{sec:geo}

\begin{figure}
    \centering
    \includegraphics[width=0.48\textwidth]{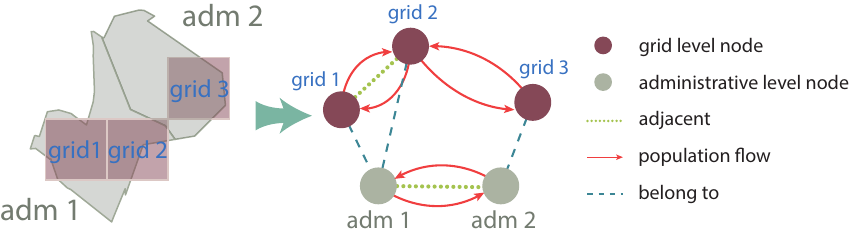}
    \caption{The illustration of the heterogeneous graph.}
    \label{fig:geospatial}
\end{figure}

Locations inherently possess spatial semantic information, inspiring the utilization of graph neural networks to learn location representation from the adjacency graph and flow graph~\cite{wang2022learning,wang2022graph,qin2023disenpoi,zhou2025disentangled}. However, these methods tend to learn embeddings independently from each graph and then merge the embeddings as the location representation, disregarding the interdependencies between the two graphs.
Towards this, we introduce a heterogeneous graph, as illustrated in Figure~\ref{fig:geospatial}, that combines both flow and adjacency relationships between locations. 
Furthermore, we expect to integrate multi-scale urban structural information into the graph. 
Considering the administrative divisions among regions, we build an admin-level graph with the administrative areas as nodes to enrich the spatial connections between locations. In this way, the heterogeneous graph considers both the grid-level and admin-level flow and adjacency relationships, along with the inclusion relationship between grids and administrative areas. The scale of administrative areas can be chosen as block, tract, town, etc.
In cases where administrative divisions are unavailable, coarser-grained grids can be used as a substitute for administrative regions.

We formulate the heterogeneous graph as $G^H = (V_{\rm grid}, V_{\rm adm}, E^f_{\rm grid}, E^f_{\rm adm}, E^a_{\rm grid}, E^a_{\rm adm}, E^i)$, where $V_{\rm grid}$ and $V_{\rm adm}$ are the node sets of grid and administrative area, $E^f_*$ and $E^a_*$ $(*\in[\rm grid, \rm adm])$ are the flow and adjacency edge sets, and $E^i$ is the inclusion edge sets. The undirected edge $e^i_{pq}, e^i_{pq} \in E^i$, means that the grid $v^{\rm grid}_p$ is located in the administrative area $v^{\rm adm}_q$. The description of edges in $E^f_*$ and $E^a_*$ remain consistent with Preliminary~\ref{sec:pre}. 
Moreover, the flow edge in our study includes hourly population flow between connected nodes. The edge is defined as $\mathbf{e}^f_{pq} = \{ f^0_{pq}, \dots, f^{23}_{pq} \}$, where $f^t_{pq}$ means the number of trips from $v_p$ to $v_q$ during the time interval from the hour $t$ to $(t+1)$, $t\in[0,23)$.

The node embeddings incorporate neighbor information from various edges to enhance the semantic richness. We employ different message-passing mechanisms based on the characteristics of the heterogeneous edges. 
To model the adjacency and inclusion edges, which primarily encode straightforward topological relationships without additional attribute information, we employ GraphSAGE~\cite{hamilton2017inductive} as the backbone for message passing. GraphSAGE’s innovative neighbor-sampling and aggregation mechanisms are particularly well-suited for handling large-scale graphs where edge semantics are simplistic and the main focus is on capturing node adjacency relationships. Furthermore, its efficiency and scalability make it an effective choice for dynamically evolving network structures embedded within urban mobility systems. The graph layer is defined as:
\begin{equation}
    \mathbf{x}^{(n)}_p = W^1 \mathbf{x}^{(n-1)}_p + W^2 {\rm mean}_{q \in N(j)}\mathbf{x}^{(n-1)}_q,
\end{equation}
where $\mathbf{x}^{(n)}$ is node embedding in the $n$-th layer, $W^1$ and $W^2$ are learnable parameters,  and ``${\rm mean}$'' is the mean operator to aggregate the neighborhood. The initial node features $\mathbf{x}^{(0)}$ comprises word embeddings of region IDs and the time-varying population count within the region. Regions refer to grids or administrative areas.

For flow edges, which capture rich semantic information such as human mobility flows and their associated attributes, we employ an edge-considered variant of the Graph Attention Network (GAT)~\cite{velickovic2017graph}. GAT leverages attention mechanisms to adaptively weight and aggregate edge features, enabling precise modeling of edge-level semantics in spatial-temporal contexts. This capacity for dynamic weighting makes GAT particularly effective for capturing nuanced relationships expressed through flow data in urban networks. This is formulated as:
\begin{equation}
    \mathbf{x}^{(n)}_p = \alpha_{pp} W^{(n-1)}_p \mathbf{x}^{(n-1)}_p + \sum_{q \in N(p)} \alpha_{pq} W^{(n-1)}_q \mathbf{x}^{(n-1)}_q,
\end{equation}
where $W^{(n-1)}_p$ and $W^{(n-1)}_q$ are learnable parameters. 
The attention coefficient $\alpha_{pq}$ is calculated as:
\begin{equation}
    \alpha_{pq} = \frac{exp(\sigma(\mathbf{a}^T [W_p\mathbf{x}_p || W_q \mathbf{x}_q || W_{pq} \mathbf{e}^f_{pq}]))}{
                    \sum_{s\in N(p) \cup \{p\}} exp(\sigma(\mathbf{a}^T [W_p\mathbf{x}_p || W_s \mathbf{x}_s || W_{ps} \mathbf{e}^f_{ps}]))},
\end{equation}
where $\sigma$ is $\rm LeakyReLU$ and these $\mathbf{x}$ are the abbreviations of $\mathbf{x}^{(n-1)}$. These $W_*, *\in\{p, q, pq, s, ps\}$ are trainble paramters.

After $n$-layers heterogeneous graph propagation, we obtain the final grid representation as the learned location embedding. These learning processes can be summarized as:
\begin{equation}
    {\rm H}^{\rm loc}_k = {\rm GraphEmbedder}({\rm C}^{\rm loc}_k).
\end{equation}

It is important to note that while GraphSAGE and GAT are employed in this framework for their suitability to specific edge types, the proposed framework is inherently modular. The GNN components can be replaced with alternative architectures as advancements in graph neural network research continue to emerge. This flexibility ensures that the framework remains adaptable and scalable, allowing integration of state-of-the-art techniques to further enhance the modeling of spatial-temporal mobility dynamics in urban systems.

\subsection{Hierarchical Trajectory Encoder}

Urban residents exhibit multi-scale temporal regularities, such as daily and weekly patterns in their travel routines, as mentioned in the Introduction~\ref{sec:intro}. To capture these periodicities more effectively, we introduce a hierarchical trajectory encoder with two layers: the daily encoder and the weekly encoder. The daily encoder captures daily travel transitions, while the weekly encoder focuses on weekly travel variation patterns. By considering information from the corresponding day a week ago and the within-week behavioral changes, our model is capable of predicting trips on the eighth day with enhanced accuracy.

Both the daily encoder and the weekly encoder utilize a positional encoding module and a Transformer encoder~\cite{vaswani2017attention}, denoted as ${\rm DailyEnc}$ and ${\rm WeeklyEnc}$, to capture sequential relationships. 
The Transformer encoder mainly consists of a multi-head self-attention mechanism and a fully connected feed-forward neural network (FNN). The attention mechanism applies various heads to simultaneously attend to different aspects, enhancing the model to capture complex dependencies from sequences and acquire information-rich representation. The detailed descriptions of the positional encoding module and the attention mechanism refer to~\cite{vaswani2017attention}.

We utilize two hierarchical encoders, namely spatial encoder and temporal encoder, to model the decoupled location chain and duration chain. With the daily encoder, we have the $k$-th daily chain representation $\mathbf{h}^{\rm loc}_k \in \mathbb{R}^{d_{hl}}$ and $\mathbf{h}^{\rm dur}_k \in \mathbb{R}^{d_{ht}}$ as follows:
\begin{equation}
    \mathbf{h}^{{\rm dur}}_k = {\rm DailyEnc^{dur}}({\rm H}^{\rm dur}_k),
\end{equation}
\begin{equation}
    \mathbf{h}^{{\rm loc}}_k = {\rm DailyEnc^{loc}}({\rm H}^{\rm loc}_k).
\end{equation}
$d_{hl}$ and $d_{ht}$ denotes the dimension of $\mathbf{h}^{\rm loc}_k$ and $\mathbf{h}^{\rm dur}_k$.
${\rm H}^{\rm loc}_k \in \mathbb{R}^{|{\rm C}^{\rm loc}_k| \times d_{el}}$ is the embedding of location chain ${\rm C}^{\rm loc}_k$ learned with the geospatial embedder, as introduced in the subsection~\ref{sec:geo}. ${\rm H}^{\rm dur}_k \in \mathbb{R}^{|{\rm C}^{\rm dur}_k| \times d_{et}}$ is the embedding of duration chain ${\rm C}^{\rm dur}_k$ learned with a word embedder. The word embedder receives the vocabulary size and embedding dimension as input, learning dense vector representations for each word. In contrast to one-hot encoding, it generates more compact and semantically informative embeddings, capturing contextual relationships between words. $d_{el}$ and $d_{et}$ denote the embedding dimension of location and duration.
Then we obtain the historical information representation $\mathbf{z}^{\rm loc} \in \mathbb{R}^{d_{zl}}$ and $\mathbf{z}^{\rm dur} \in \mathbb{R}^{d_{zt}}$ through the weekly encoder, formulated as:
\begin{equation}
    \mathbf{z}^{\rm dur} = {\rm WeeklyEnc^{dur}}(\{ \mathbf{h}^{\rm dur}_{k-6}\oplus\mathbf{h}^{\rm loc}_{k-6},\dots, \mathbf{h}^{\rm dur}_k\oplus\mathbf{h}^{\rm loc}_k \}),
\end{equation}
\begin{equation}
    \mathbf{z}^{\rm loc} = {\rm WeeklyEnc^{loc}}(\{ \mathbf{h}^{\rm loc}_{k-6},\dots, \mathbf{h}^{\rm loc}_k \}).
\end{equation}
$d_{zl}$ and $d_{zt}$ denotes the dimension of $\mathbf{z}^{\rm loc}$ and $\mathbf{z}^{\rm dur}$.
Here, the symbol $\oplus$ represents vector concatenation and the concatenated vector, $\mathbf{h}^{\rm dur}_k\oplus\mathbf{h}^{\rm loc}_k$, has the dimension $d^{\rm loc} + d^{\rm dur}$.
The missing days during a week are masked in the weekly encoder to preserve the temporal structure integrity.


\subsection{Trajectory Decoder}

The trajectory decoder is comprised of a spatial decoder and a temporal decoder. Considering the dependency between duration and location, we expect to predict the location chain first and then utilize it to enhance the prediction of the duration chain. Inspired by the natural language process tasks, we add the start token $\rm <SOS>$ and end token $\rm <EOS>$ into the location chain to address the challenges of the unknown start location and the uncertain chain length. Besides, we employ the Transformer decoder~\cite{vaswani2017attention} as the backbone of the spatial decoder, along with a positional encoding module and an FNN. The self-attention mechanism in the Transformer decoder and positional encodings enable the model to effectively capture long-range dependencies and maintain the temporal order of the input sequence, thereby alleviating accumulative errors in multi-step prediction tasks.
Commencing from the start token, the spatial decoder utilizes the historical information and predicted results to make iterative predictions for the daily location chain. This process is formulated as:
\begin{equation}
    \hat{l}_{i+1} = {\rm SpatialDec}(\{s_0, \hat{s}_1 \dots, \hat{s}_{i} \}, z^{\rm loc}),
\end{equation}
where $\rm SpatialEnc$ is the abbreviation of the spatial decoder, $\hat{s}_i$ is the embedding of $i$-th predicted location $\hat{s}_i = {\rm GraphEmbed} (\hat{l}_i)$ and $s_0$ is the embedding of the start token. The iterative prediction process halts upon the prediction of the end token. Consequently, the predicted location chain is $\hat{\rm C}^{\rm loc}= \{ \hat{l}_1 \dots, \hat{l}_{m} \}$, with $m$ representing the number of location appeared on the target day.

The temporal decoder comprises a positional encoding module, Transformer encoder, and FNN. We utilize the Transformer encoder to enable the model to employ attention mechanisms for observing daily location access patterns when predicting duration. Historical time data and predicted location representations are combined into the time encoder to derive the forecasted duration chain, formulated as:

\begin{equation}
    \resizebox{0.9\hsize}{!}{$\begin{aligned}
        \hat{\rm C}^{\rm dur} = \{ \hat{t}_1 \dots, \hat{t}_{m} \} = {\rm TemporalDec}(\{ \hat{s'}_1 \oplus z^{\rm dur} , \dots, \hat{s'}_m \oplus z^{\rm dur} \}),
       \end{aligned}$}
\end{equation}

where ${\rm TemporalDec}$ is the abbreviation of the temporal decoder and $\oplus$ represents vector concatenation as mentioned before. $\hat{s'}_k = {\rm FNN}(\hat{s}_k)$, where a FNN is used to adjust the dimension of the predicted location embeddings.  
Finally, we combine the predicted duration chain and location chain to derive the predicted trajectory for the following day.

\subsection{Model Optimization}

We apply the most commonly used cross-entropy loss for the location chain prediction. The Huber loss is used for the duration chain prediction as it integrates the advantages of MSE and MAE. The total loss function is defined as:
\begin{equation}
    L = L_{ce}(l_j, \hat{l}_j) + \lambda \cdot L_{huber}(t_j, \hat{t}_j),
\end{equation}
where $ L_{huber}(t_j, \hat{t}_j) = 
        \begin{cases}
            (t_j - \hat{t}_j)^2, |t_j - \hat{t}_j| < 1 \\
            |t_j - \hat{t}_j| - 1/2, otherwise
        \end{cases}$
, $L_{ce}(l_j, \hat{l}_j) = - {l_j}\log{\hat{l}_j}$, and $\lambda$ is a hyperparamter to adjust loss weight. 
$l_j$ and $t_j$ are the targets, while $\hat{l}_j$ and $\hat{t}_j$ are the predicted values.

\section{Experiments}

\subsection{Experimental Settings}

\subsubsection{Data Description} 
Experiments are conducted on four private~\footnote{The dataset is not publicly disclosed due to privacy constraints.} urban-scale datasets, \textit{Boston}, \textit{Los Angeles}~(LA), \textit{San Francisco Bay Area}~(SFBay), \textit{Shanghai}, collected from real-world mobile records, and one public dataset \textit{Tokyo}~\cite{yabe2023metropolitan}, comprising synthetic mobility data. The preprocessing steps for these four CDR datasets follow~\cite{jiang2016timegeo}. Besides, we partition each city into uniform grids with grid sizes of 1 kilometer. As described in Preliminary, a day is divided into multiple time windows, and only the record with the longest stay duration is retained within each window. Boston, LA, and SFBay utilize half-hour windows, whereas Shanghai and Tokyo use hour-long windows. 

The Tokyo dataset, as distinct from the other four datasets, undergoes two specific preprocessing steps. First, we identified users' home locations by leveraging the approach in~\cite{jiang2016timegeo}, retaining only those users whose home location could be determined, thus focusing on the resident population. A location is classified as ``home" if it accounts for more than 30\% of visits during nighttime hours (10PM–6AM). Second, we applied mobility behavior constraints inspired by~\cite{schneider2013unravelling,jiang2016timegeo}, ensuring that 90\% of a user’s daily trajectories include fewer than 10 unique locations.

The statistics of the datasets are presented in Table~\ref{tab:data}, where ``\#~Days" means the average number of days and ``\#~Admins" means the number of the administrative division. \textit{Boston}, \textit{LA}, and \textit{SFBay} employ census tracts as administrative units, whereas Shanghai uses towns. In the Tokyo dataset, due to blurred exact coordinates, we partitioned administrative regions using 7x7 grid squares. The training, validation, and test sets are obtained by partitioning the original dataset in a ratio of 6:1:3 based on the time span.

\begin{table}[h]
\caption{Statistics of the preprocessed urban-scale datasets.}
\label{tab:data}
    \centering
    \resizebox{\linewidth}{!}{
    \begin{tabular}{c|cccccc}
        \toprule
        & Boston & LA & SFBay & Shanghai & Tokyo \\
        \midrule
        \# Individuals & 50,000 & 50,000 & 50,000 & 20,000 & 16,151 \\
        \# Locations & 10,294 & 6,188 & 8,032 & 16,050 & 7,153 \\
        \# Admins & 1,126 & 2,493 & 1,406 & 234 & 225 \\
        \# Records & 6,749,540 & 7,768,373 & 7,700,986 & 8,209,910 & 7,502,839 \\
        \# Days & 39 & 41 & 41 & 99 & 75 \\
        Time range & \makecell{Feb. 20, 2010\\ to \\Mar. 30, 2010} 
                   & \makecell{Oct. 15, 2012\\ to \\Nov. 24, 2012}  
                   & \makecell{Oct. 15, 2012\\ to \\Nov. 24, 2012} 
                   & \makecell{Jan. 1, 2014\\ to  \\Apr. 29, 2014} 
                   & Unknown \\
        \bottomrule
    \end{tabular}
    } 
\end{table}

\subsubsection{Baselines}

We compared 11 baseline models, which can be categorized into three groups based on their task orientation: next-location prediction models, successive location prediction models, and mid-term mobility prediction models. A brief overview of these baseline models is provided below.

Next-location prediction models
\begin{itemize}
    \item \textbf{CFPRec}~\cite{zhang2022next} develops a two-layer attention mechanism where future time embeddings are used to query the most valuable historical information.
    \item \textbf{GETNext}~\cite{yang2022getnext} utilizes a flow map to learn location representation and a Transformer to encode historical sequences and make predictions.
    \item \textbf{GFlash}~\cite{rao2022graph}, originally named Graph-Flashback, designs a knowledge graph to learn the transition probabilities between locations and employs GNNs to learn location representations in a homogeneous graph.
    \item \textbf{HGARN}~\cite{tang2022hgarn} introduces a hierarchical graph to extract dependencies among activities and locations. The location representations are learned with graph attention networks.
    \item \textbf{SNPM}~\cite{yin2023next} applies a dynamic graph method to discover neighbors of current locations and learn the location representation, along with an attention-based structure to make predictions.
    \item \textbf{AGRAN}~\cite{wang2023adaptive} proposes an adaptive graph learning module to learn an optimized topology of location. An attention-based network is adopted to predict locations.
    \item \textbf{MTNet}~\cite{huang2024learning} construct a tree structure from individual trajectories to capture the preferences in multi-granularity time slots.
\end{itemize}

Successive prediction model
\begin{itemize}
    \item \textbf{RLMob}~\cite{luo2022rlmob} models the successive mobility prediction as Markov Decision Process and proposes an actor-critic framework with an instance of Proximal Policy Optimization (PPO).
\end{itemize}

Mid-term mobility prediction model
\begin{itemize}
    \item \textbf{HTAED}~\cite{li2020hierarchical} introduces an LSTM-based encoder-decoder model with a hierarchical temporal attention module for multi-day mobility prediction.
\end{itemize}

\subsubsection{Evaluation Metrics}
We evaluate these models with four metrics: \textit{Acc}, \textit{DevDist}, \textit{TravelDist}, and \textit{DepartTime}. \textit{Acc} indicates the proportion of the accurately predicted locations. \textit{DevDist} is the average deviation distance, measured in kilometers,  between all actual and predicted locations. \textit{TravelDist} and \textit{DepartTime} quantify the Jensen-Shannon divergence (JSD) between actual and predicted travel distances and departure times for individuals. 

\textit{Acc} and \textit{DevDist} are assessed point-wise, while \textit{TravelDist} and \textit{DepartTime} are evaluated individually. Specifically, \textit{Acc} describes the proportion of the accurately predicted locations, calculated with $Acc = \frac{Count(\hat{l} == l)}{N_l}$, where $Count()$ measures the number of correctly predicted locations, with $N_l$ denoting the total number of locations in the trajectory. \textit{DevDist} measures the averaged deviation distance between actual and predicted locations, formulated with $DevDist = mean_{N_l} \Big ( \big ( distance(\hat{l},l) \big ) \Big ) $. $distance()$ calculates the distance between two locations in kilometers, whereas $mean_{N_l}()$ computes the average over $N_l$ locations. \textit{TravelDist} initially computes the distribution of individual travel distances, evaluates the Jensen-Shannon divergence (JSD) between the actual and predicted distributions per individual, and subsequently averages these differences across all users. It is formally defined by $mean_{N_l} \big ( JSD ( \hat{p}_{dist}, p_{dist} ) \big )$, where $\hat{p}_{dist}$ and $p_{dist}$ denote the predicted and actual distribution of travel distances. The calculation of \textit{DepartTime} is similar to \textit{TravelDist}, focusing on the distribution of departure times. Its formal definition is $mean_{N_l} \big ( JSD ( \hat{p}_{time}, p_{time} ) \big )$, where $\hat{p}_{time}$ and $p_{time}$ denote the predicted and actual distribution of departure time.

The improved performance is reflected in larger \textit{Acc}, as well as smaller \textit{DevDist},  \textit{TravelDist}, and \textit{DepartTime}.

\subsubsection{Implementation}
These baselines are implemented with their open-source codes. Hyperparameter tuning is performed on the validation set to attain optimal performance. The implementation of MSTDP is based on the PyTorch framework. The heterogeneous GNNs adopt a two-layer structure, while the layer numbers of Transformer encoder and decoder are also set to 2. The embedding dimensions of location and time are 512 and 512. The hidden dimensions of location and time are 1024 and 512. The head numbers of attention in the Transformer encoder and decoder are set to 8.  
We apply the Adam optimizer to train MSTDP with a learning rate of $10^{-4}$. The hyperparameter $\lambda$ is set to 1. The number of training epochs is 60. 
Our source code is available at https://github.com/urbanmobility/MSTDP.git

\subsubsection{Task Settings}
We evaluate models in two tasks: next day prediction and next week prediction. For next day prediction, all models iteratively forecast the location of the next T timeslots by considering historical information and previously predicted results. This means that the prediction of the $i$-th location $\hat{l}^i_r$ is based on the true historical information $\{{\rm D}_0, \dots, {\rm D}_{k} \}$ and the predicted locations of the preceding i-1 points $\{ \hat{l}^0_p, \dots, \hat{l}^{i-1}_q \}$, where $i \in [1,{\rm T}]$. The one-hour time window corresponds to T=24, while the half-hour time window corresponds to T=48. Similarly, for next week prediction, the next $7*{\rm T}$ locations in the following week are also iteratively forecasted. That is, $\hat{\rm D}_{k+i}$ is predicted with $\{{\rm D}_0, \dots, {\rm D}_{k} \}$ and $\{\hat{\rm D}_{k+1}, \dots, \hat{\rm D}_{k+i-1} \}$, where $i \in [2,7]$.

\begin{table*}[ht]
\caption{Performance comparison of all methods on three real-world datasets. The optimal and second-best results in each column are highlighted with bold text and underlining.}
\label{tab:metrics}
    \centering
    \resizebox{\linewidth}{!}{
    \begin{tabular}{c|c|c|cccccccccc}
        \toprule
            & & & CFPRec & GETNext & GFlash & HGARN & SNPM & AGRAN & MTNet & HTAED & RLMob & \textbf{MSTDP} \\
        \midrule
            \multirow[c]{8}{*}{\rotatebox{90}{Boston}} & \multirow[c]{4}{*}{\rotatebox{90}{Day}} & Acc $\uparrow$
                & 0.420 & 0.323 & 0.360 & 0.377 & 0.268 & 0.496 & 0.498 & \underline{0.564} & 0.363 & \textbf{0.584} \\
            & & DevDist $\downarrow$
                & 4.991 & 5.869 & 6.406 & 5.855 & 14.166 & 5.506 & 7.969 & \underline{4.114} & 7.744 & \textbf{3.676} \\
            & & TravelDist $\downarrow$
                & 0.655 & 0.675 & 0.667 & 0.621 & 0.619 & 0.635 & 0.650 & \underline{0.509} & 0.606 & \textbf{0.361} \\
            & & DepartTime $\downarrow$
                & 0.541 & 0.656 & 0.570 & 0.656 & 0.654 & 0.647 & 0.658 & \underline{0.455} & 0.636 & \textbf{0.419} \\
            \cmidrule(r){2-13}
            & \multirow[c]{4}{*}{\rotatebox{90}{Week}} & Acc $\uparrow$
                & 0.340 & 0.322 & 0.351 & 0.354 & 0.264 & 0.371 & 0.433 & \underline{0.444} & 0.322 & \textbf{0.470} \\
            & & DevDist $\downarrow$
                & 7.470 & 5.916 & 6.471 & 6.603 & 14.220 & 6.123 & 8.357 & \underline{5.756} & 8.236 & \textbf{5.169} \\
            & & TravelDist $\downarrow$
                & 0.652 & 0.672 & 0.659 & 0.627 & 0.615 & 0.613 & 0.671 & \underline{0.583} & 0.681 & \textbf{0.425} \\
            & & DepartTime $\downarrow$
                & 0.571 & 0.657 & 0.653 & 0.604 & 0.656 & 0.635 & 0.669 & \underline{0.530} & 0.668 & \textbf{0.418} \\
        \midrule
            \multirow[c]{8}{*}{\rotatebox{90}{Shanghai}} & \multirow[c]{4}{*}{\rotatebox{90}{Day}} & Acc $\uparrow$
                & 0.557 & 0.505 & 0.565 & 0.569 & 0.563 & 0.582 & 0.540 & \underline{0.602} & 0.585 & \textbf{0.617} \\
            & & DevDist $\downarrow$
                & 3.591 & 8.227 & 4.255 & 3.206 & 3.202 & 3.148 & 6.064 & \underline{3.064} & 3.193 & \textbf{2.729} \\
            & & TravelDist $\downarrow$
                & 0.578 & 0.596 & 0.596 & 0.591 & 0.537 & 0.506 & 0.630 & \underline{0.436} & 0.650 & \textbf{0.325} \\
            & & DepartTime $\downarrow$
                & 0.505 & 0.566 & 0.573 & 0.561 & 0.566 & 0.476 & 0.593 & \underline{0.297} & 0.597 & \textbf{0.197} \\
            \cmidrule(r){2-13}
            & \multirow[c]{4}{*}{\rotatebox{90}{Week}} & Acc $\uparrow$
                & 0.463 & 0.469 & 0.518 & 0.561 & 0.561 & 0.539 & 0.523 & \underline{0.566} & 0.521 & \textbf{0.612} \\
            & & DevDist $\downarrow$
                & 8.989 & 11.880 & 8.694 & 3.482 & \underline{3.224} & 3.477 & 7.749 & 3.456 & 7.301 & \textbf{2.780} \\
            & & TravelDist $\downarrow$
                & 0.489 & 0.496 & 0.494 & 0.598 & 0.539 & 0.568 & 0.569 & \underline{0.484} & 0.608 & \textbf{0.335} \\
            & & DepartTime $\downarrow$
                & 0.408 & 0.522 & 0.527 & 0.571 & 0.566 & 0.534 & 0.549 & \underline{0.383} & 0.559 & \textbf{0.227} \\
        \midrule
            \multirow[c]{8}{*}{\rotatebox{90}{Tokyo}} & \multirow[c]{4}{*}{\rotatebox{90}{Day}} & Acc $\uparrow$
                & 0.566 & 0.580 & 0.505 & 0.483 & 0.576 & 0.584 & 0.564 & \underline{0.587} & 0.492 & \textbf{0.595} \\
            & & DevDist $\downarrow$
                & 2.295 & 2.099 & 2.666 & 2.497 & 2.466 & \underline{2.010} & 3.012 & 2.039 & 3.758 & \textbf{1.916} \\
            & & TravelDist $\downarrow$
                & 0.587 & 0.622 & 0.592 & 0.627 & 0.607 & 0.599 & 0.600 & \underline{0.567} & 0.650 & \textbf{0.539} \\
            & & DepartTime $\downarrow$
                & 0.493 & 0.618 & 0.344 & 0.317 & 0.610 & 0.557 & 0.601 & \underline{0.275} & 0.604 & \textbf{0.108} \\
            \cmidrule(r){2-13}
            & \multirow[c]{4}{*}{\rotatebox{90}{Week}} & Acc $\uparrow$
                & 0.472 & 0.568 & 0.504 & 0.457 & 0.565 & \underline{0.571} & 0.559 & \underline{0.571} & 0.486 & \textbf{0.588} \\
            & & DevDist $\downarrow$
                & 3.918 & 2.092 & 2.672 & 2.589 & 2.478 & \underline{2.051} & 3.254 & 2.450 & 3.439 & \textbf{1.997} \\
            & & TravelDist $\downarrow$
                & \underline{0.549} & 0.627 & 0.592 & 0.623 & 0.608 & 0.577 & 0.622 & 0.574 & 0.652 & \textbf{0.547} \\
            & & DepartTime $\downarrow$
                & 0.416 & 0.620 & 0.343 & 0.303 & 0.610 & 0.610 & 0.610 & \underline{0.297} & 0.602 & \textbf{0.120} \\
        \bottomrule
    \end{tabular}
    }
\end{table*}

\begin{table}[ht]
\caption{Metric comparison on \textit{LA} and \textit{SFBay} datasets.}
\label{tab:metric2}
    \centering
    \resizebox{\linewidth}{!}{
    \begin{tabular}{c|c|c|ccccc}
        \toprule
            & & & AGRAN & MTNet & HTAED & RLMob & \textbf{MSTDP} \\
        \midrule
            \multirow[c]{8}{*}{\rotatebox{90}{LA}} & \multirow[c]{4}{*}{\rotatebox{90}{Day}} & Acc $\uparrow$
                & 0.485 & 0.462 & \underline{0.550} & 0.354 & \textbf{0.573} \\
            & & DevDist $\downarrow$
                & 4.982 & 5.005 & \underline{4.194} & 6.412 & \textbf{3.818} \\
            & & TravelDist $\downarrow$
                & 0.648 & 0.684 & \underline{0.547} & 0.686 & \textbf{0.422} \\
            & & DepartTime $\downarrow$
                & 0.661 & 0.667 & \underline{0.461} & 0.674 & \textbf{0.374} \\
            \cmidrule(r){2-8}
            & \multirow[c]{4}{*}{\rotatebox{90}{Week}} & Acc $\uparrow$
                & 0.424 & 0.412 & \underline{0.445} & 0.338 & \textbf{0.460} \\
            & & DevDist $\downarrow$
                & 5.304 & 5.492 & \underline{5.130} & 6.542 & \textbf{4.921} \\
            & & TravelDist $\downarrow$
                & 0.656 & 0.674 & \underline{0.592} & 0.685 & \textbf{0.477} \\
            & & DepartTime $\downarrow$
                & 0.673 & 0.673 & \underline{0.546} & 0.677 & \textbf{0.392} \\
        \midrule
            \multirow[c]{8}{*}{\rotatebox{90}{SFBay}} & \multirow[c]{4}{*}{\rotatebox{90}{Day}} & Acc $\uparrow$
                & 0.452 & 0.442 & \underline{0.529} & 0.324 & \textbf{0.558} \\
            & & DevDist $\downarrow$
                & 6.818 & 7.210 & \underline{5.784} & 10.576 & \textbf{5.222} \\
            & & TravelDist $\downarrow$
                & 0.645 & 0.670 & \underline{0.562} & 0.679 & \textbf{0.436} \\
            & & DepartTime $\downarrow$
                & 0.587 & 0.666 & \underline{0.461} & 0.673 & \textbf{0.372} \\
            \cmidrule(r){2-8}
            & \multirow[c]{4}{*}{\rotatebox{90}{Week}} & Acc $\uparrow$
                & 0.399 & 0.404 & \underline{0.416} & 0.297 & \textbf{0.430} \\
            & & DevDist $\downarrow$
                & 7.660 & 7.883 & \underline{7.169} & 10.706 & \textbf{6.945} \\
            & & TravelDist $\downarrow$
                & 0.664 & 0.672 & \underline{0.609} & 0.684 & \textbf{0.493} \\
            & & DepartTime $\downarrow$
                & 0.598 & 0.675 & \underline{0.552} & 0.680 & \textbf{0.384} \\
        \bottomrule
    \end{tabular}
    }
\end{table}

\subsection{Metric Comparison}
We initially evaluate all models using datasets \textit{Boston}, \textit{Shanghai}, and \textit{Tokyo}, and subsequently select the more recent baselines to compare against datasets \textit{LA} and \textit{SFBay}. The results are exhibited in Table~\ref{tab:metrics} and Table~\ref{tab:metric2}.
From these results, we draw the following observations and analysis: (1) MSTDP demonstrates superior performance on both datasets, validating its effectiveness in mid-term mobility prediction tasks. Compared to the top-performing baseline, MSTDP exhibits improvements of 3.6\% in \textit{Acc}, 9.0\% in \textit{DevDist}, 20.3\% in \textit{TravelDist} and 19.4\% in \textit{DepartTime}. A smaller \textit{DevDist} of MSTDP indicates closer proximity between the predicted and the actual locations. Furthermore, lower metrics for \textit{TravelDist} and \textit{DepartTime} demonstrate improved individual consistency between predicted and actual trajectories regarding travel time and distance patterns in our model.
(2) Among these baselines, HTAED excels due to its dual-layer attention mechanism tailored for successive prediction, effectively weighing intra-day and inter-day behavioral patterns. In contrast, RLMob, which employs a reinforcement learning framework for successive prediction, falls short by relying solely on RNNs to model transition relationships, resulting in less favorable performances. 
These baselines, initially developed for next-location prediction, overlook temporal patterns such as daily and weekly cycles, which complicates their direct application to mid-term forecasting tasks. Moreover, when applied to \textit{Boston}, \textit{LA}, or \textit{SFBay} datasets, these baseline predictions show significant performance degradation compared to \textit{Shanghai} and \textit{Tokyo}, notably highlighted by SNPM. This discrepancy primarily stems from varying time window sizes across datasets; for instance, \textit{Boston} employs a half-hour window, leading to increased location redundancy and longer daily trajectories. These factors present substantial challenges for model performance, as discussed in Introduction~\ref{sec:intro}.


\subsection{Statistical Physics Analysis}
The metric comparison in Table~\ref{tab:metrics} does not sufficiently portray the model's performance. Therefore, we conduct a comprehensive statistical physics analysis on the \textit{Boston} dataset from multiple perspectives to thoroughly examine the performance of these models.

\begin{figure}[ht]
    \centering
    \includegraphics[width=0.48\textwidth]{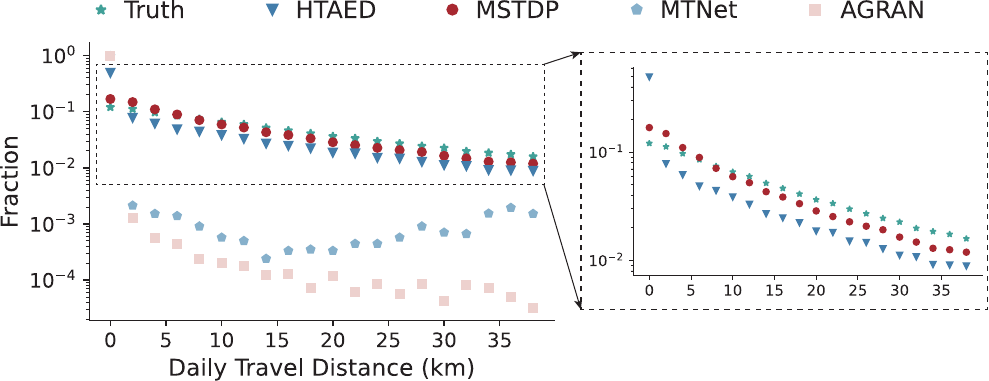}
    \caption{Daily travel distances. }
    \label{fig:travelDist}
\end{figure}

\textbf{Daily travel distance.} 
Figure~\ref{fig:travelDist} presents the statistical distribution of daily travel distances for individuals.
These baselines predict a higher number of zero travel distances. It indicates that the models predict a greater occurrence of full-day stays at the same location, which we refer to as ``full-day stays''. Unlike these baselines, MSTDP demonstrates closer alignment between predicted daily travel distances and actual distributions.

\begin{figure}[ht]
    \centering
    \includegraphics[width=0.48\textwidth]{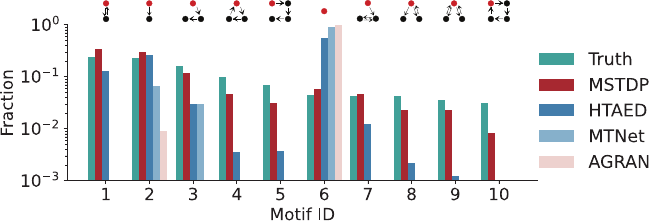}
    \caption{Top 10 daily motifs. }
    \label{fig:motif}
\end{figure}

\textbf{Human mobility motif.} Motif describes people's mobility pattern during one day with abstract network~\cite{schneider2013unravelling}. 
Figure~\ref{fig:motif} illustrates the ten most prevalent motifs in actual daily trajectories, covering 74.6\% of the population. MTNet and AGRAN primarily identify three patterns, with a notable emphasis on pattern ID=6, indicative of their focus on daily full-day stays. Our model adeptly captures these prevalent travel patterns, demonstrating performance superiority over HTAED through close alignment with the actual distribution.

\begin{figure}
    \centering
    \includegraphics[width=0.48\textwidth]{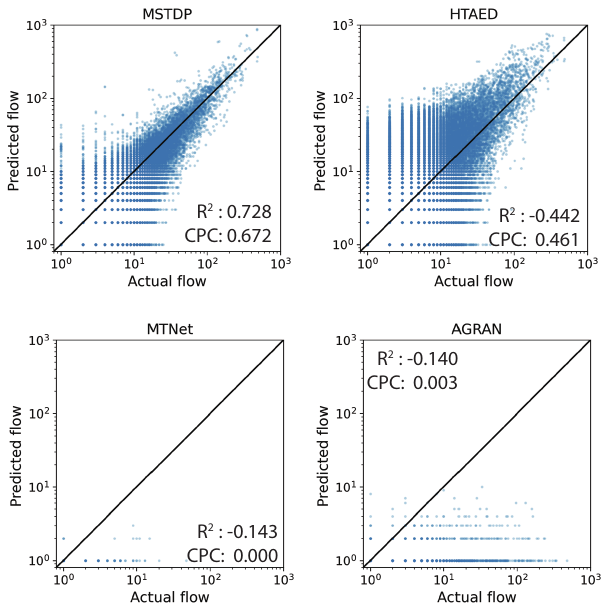}
    \caption{Origin-Destination flow pairs between actual and predicted flows.}
    \label{fig:flowScatter}
\end{figure}

\textbf{Origin-Destination (OD) flow pairs.} We collect the grid-level OD flow pairs between the actual and predicted flows and illustrate them in Figure~\ref{fig:flowScatter}. We calculate $\rm{R}^2$ and CPC~\cite{lenormand2016systematic} between actual and predicted flows for numerical comparison. AGRAN and MTNet predict a limited number of flows, consistent with previous analyses indicating their preference for predicting full-day stays. HTAED overestimates flows with an $\rm R^2$ of -0.442 and $\rm CPC$ of 0.461, whereas our MSTDP model achieves closer predictions to actual flow quantities, with an improved $\rm R^2$ of 0.728 and $\rm CPC$ of 0.672.

\begin{figure}
    \centering
    \includegraphics[width=0.48\textwidth]{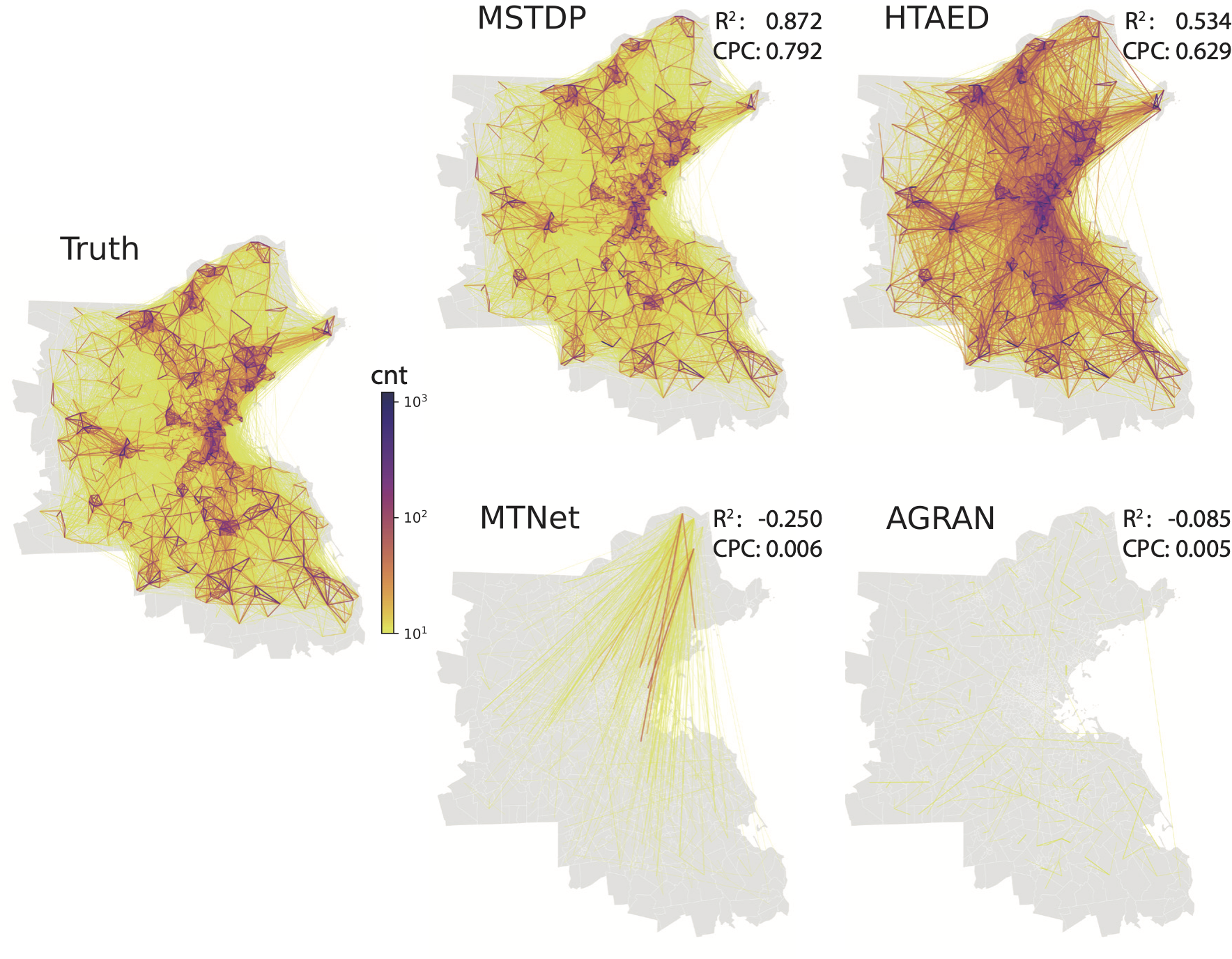}
    \caption{Origin-Destination flows in tract level.}
    \label{fig:flow}
\end{figure}

\begin{figure}[ht]
    \centering
    \includegraphics[width=0.48\textwidth]{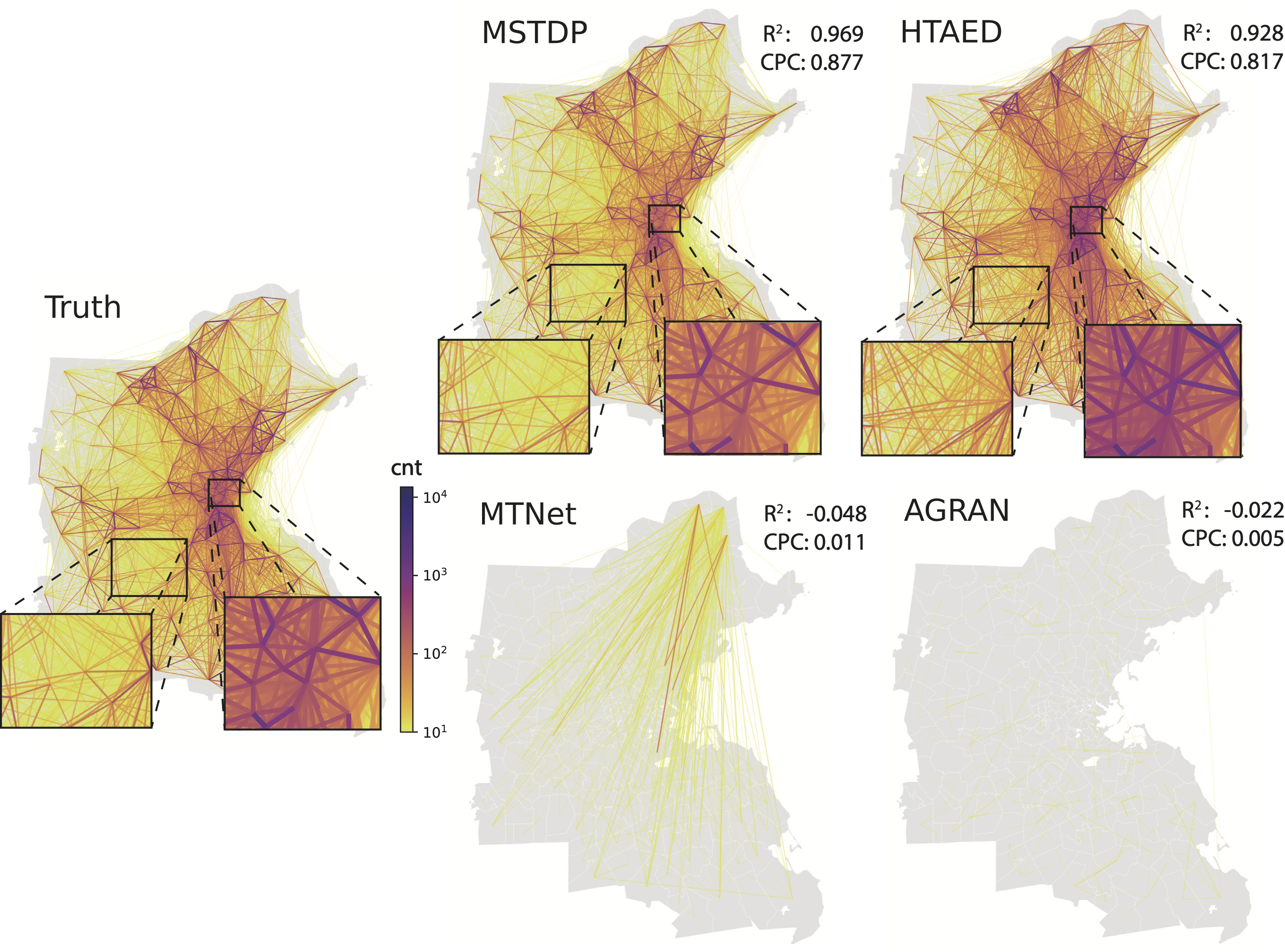}
    \caption{Origin-Destination flows in zip code level.}
    \label{fig:apdx_flow}
\end{figure}

\textbf{Origin-Destination flows in map.} Figure~\ref{fig:flow} illustrates the actual and predicted OD flows in the 
map, where the trips between grids are aggregated at the census tract level. The global travel patterns predicted by the three baseline models deviate significantly from the actual results. HTAED achieves $\rm R^2$ and CPC values of 0.534 and 0.629, whereas MSTDP achieves 0.872 and 0.792, indicating performance improvements of 63.3\% and 25.9\% in our model. In this subsection, we examine OD flows at the grid and tract levels. Additionally, Figure~\ref{fig:apdx_flow} illustrates OD flows at the zip code level. MSTDP achieves R-squared ($R^2$) and CPC values of 0.969 and 0.877, while HTAED obtains 0.928 and 0.817. The performance improvements are 4.4\% and 7.3\%, respectively. This observation suggests that larger spatial aggregations may obscure differences between models.

\begin{figure}[h]
    \centering
    \setlength{\belowcaptionskip}{-1mm}
    \includegraphics[width=0.48\textwidth]{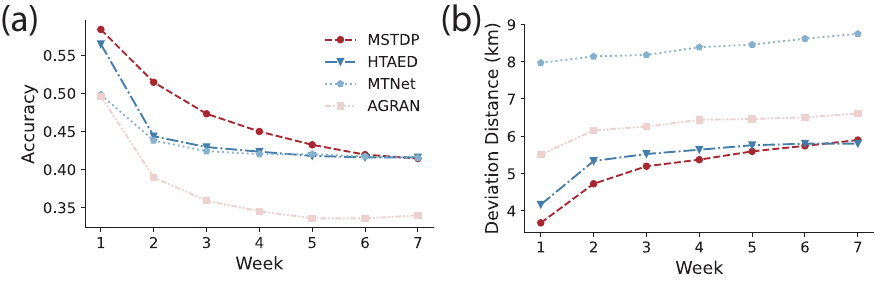}
    \caption{Metrics of consecutive seven-day forecasts. (a) Accuracy; (b) Deviation distance.}
    \label{fig:weekMetric}
\end{figure}

\textbf{Seven-day consecutive forecasting.} We employ an iterative forecasting approach to predict future travel continuously throughout the upcoming week. The accuracy and deviation distance fluctuation of the predictions over seven days are illustrated in Figure~\ref{fig:weekMetric}. As the forecasting horizon extends, the accuracy of all models gradually diminishes, alongside increased deviation distances between actual and predicted locations. This trend highlights the cumulative error inherent in iterative prediction approaches. Notably, when forecasting the subsequent second and third days, MSTDP exhibits slower performance degradation than HTAED. This disparity can be attributed to our model's broader contextual awareness, leveraging 5 and 4 days of historical data for respective predictions.

\subsection{Ablation Study}
~\label{apdx:ablat}
We compare MSTDP with three variants to examine the effectiveness of the modules in MSTDP. The ablated variants of MSTDP are as follows: (1) MSTDP-D: it eliminates the spatial-temporal decoupler and directly encodes the long trajectories; (2) MSTDP-H: it abandons the hierarchical structure and solely utilizes the daily encoder; (3) MSTDP-G: it replaces the graph-based location representation with a word embedder; (4) MSTDP-LD: the temporal decoder excludes the concatenation from the location decoder and uses a Transformer decoder structure rather than the Transformer encoder. 

The results in Table~\ref{tab:ablation} demonstrate that MSTDP outperforms these ablation variants, highlighting the effectiveness of each designed module. Among these variants, the MSTDP-H exhibits the weakest performance in both prediction tasks, likely due to reduced capability in capturing long-term dependencies and weekly periodicity following the removal of its hierarchical structure. MSTDP-LD exhibits inferior performance compared to MSTDP-D and MSTDP-G in next day prediction tasks, yet surpasses them in next week prediction tasks. This could be attributed to the incorporation of predicted locations in the temporal decoder, resulting in increased cumulative errors in next week predictions. Inaccuracies in location predictions adversely impact the model's overall performance.


\begin{table}[ht]
\caption{Ablation studies on \textit{Boston} dataset.}
\label{tab:ablation}
    \centering
    \resizebox{\linewidth}{!}{
    \begin{tabular}{c|c|cccc}
    \toprule
          & Metrics & 
                \textit{Acc} $\uparrow$ & \textit{DevDist} $\downarrow$ & \textit{TravelDist} $\downarrow$ & \textit{TravelTime} $\downarrow$ \\
    \midrule
        \multirow[c]{5}{*}{\rotatebox{90}{\makecell{Next Day \\ Prediction}}}  
         & MSTDP-D    & 0.571 & 3.891 & 0.539 & 0.560  \\
         & MSTDP-H    & 0.448 & 5.003 & 0.426 & 0.450  \\
         & MSTDP-G    & 0.575 & 3.749 & 0.381 & 0.447 \\
         & MSTDP-LD   & 0.564 & 3.870 & 0.394 & 0.424 \\
         & MSTDP      & \textbf{0.584} & \textbf{3.676} & \textbf{0.361} & \textbf{0.419}\\
    \midrule 
        \multirow[c]{5}{*}{\rotatebox{90}{\makecell{Next Week \\ Prediction}}}
         & MSTDP-D    & 0.404 & 5.634 & 0.564 & 0.570 \\
         & MSTDP-H    & 0.383 & 5.747 & 0.472 & 0.464 \\
         & MSTDP-G    & 0.405 & 5.518 & 0.461 & 0.449 \\
         & MSTDP-LD   & 0.409 & 5.375 & 0.470 & 0.440 \\
         & MSTDP      & \textbf{0.413} & \textbf{5.369} & \textbf{0.455} & \textbf{0.438} \\
    \bottomrule
    \end{tabular}
    }
    
\end{table}

\subsection{Hyperparameter Analysis}
\label{apdx:hyper}
To examine MSTDP's sensitivity to hyperparameters, we vary four hyperparameters to observe performance fluctuations: $\lambda$ (the weight of the temporal loss), head number ($\#Head$) in the Transformer encoder, embedding dimensions (Dim), and layer depths ($\#Layer$). 
For $\lambda$, we test values of 0.01, 0.1, 1, 10, and 100, which impact the optimization balance between spatial and temporal losses. For head number, we evaluate 2, 4, and 8 attention heads. Additionally, we investigate the effect of embedding dimensions, testing values of 256, 512, and 768, and layer depths, comparing configurations with 1, 2, and 3 layers. The default parameter settings are $\lambda=1$, $\#Head=8$, $Dim=512$, and $\#Layer=2$, with training epochs at 40.

Table~\ref{tab:hyper} presents the sensitivity analysis of MSTDP under various hyperparameter configurations, revealing several key trends. For the temporal loss weight ($\lambda$), larger values consistently yield better results. $\lambda=10$ achieves the highest \textit{Acc}, while $\lambda=100$ improves \textit{TravelTime}. Smaller values, such as $\lambda=0.01$ and $\lambda=0.1$, lead to lower accuracy and higher deviation distance, emphasizing the importance of effectively balancing the temporal loss. Increasing the attention head number ($\#Head$) improves performance, with the default setting ($\#Head=8$) achieving the lowest \textit{DevDist} and \textit{TravelDist}, demonstrating the advantage of richer attention mechanisms. For layer depth ($\#Layer$) and embedding dimensions ($Dim$), the default configurations of 2 layers and $Dim=512$ achieve the best overall balance. While deeper or wider models might benefit from extended training (more epochs), they also pose a heightened risk of overfitting. The default configuration achieves optimal performance across all metrics, highlighting the importance of a carefully tuned model for mid-term mobility prediction tasks.

\begin{table}[ht]
\caption{Hyperparameter analysis on \textit{Boston} dataset.}
\label{tab:hyper}
    \centering
    \resizebox{\linewidth}{!}{
    \begin{tabular}{c|c|cccc}
    \toprule
            & & \textit{Acc} $\uparrow$ & \textit{DevDist} $\downarrow$ & \textit{TravelDist} $\downarrow$ & \textit{TravelTime} $\downarrow$ \\
    \midrule
        \multirow[c]{7}{*}{\rotatebox{90}{\makecell{Next Day \\ Prediction}}}  
        & $\lambda = 0.01$     & 0.567 & 3.826 & 0.387 & 0.425     \\
        & $\lambda = 0.1$      & 0.569 & 3.825 & 0.395 & 0.430     \\
        & $\lambda = 10$       & \textbf{0.574} & 3.777 & 0.386 & 0.428     \\
        & $\lambda = 100$      & 0.571 & 3.786 & 0.400 & \textbf{0.421}     \\
        & $\# Head = 2$        & 0.568 & 3.826 & 0.392 & 0.426     \\
        & $\# Head = 4$        & 0.571 & 3.803 & 0.389 & 0.430      \\
        & $\# Layer = 1$       & 0.562 & 3.875 & 0.395 & 0.428     \\
        & $\# Layer = 3$       & \textbf{0.574} & 3.796 & 0.396 & 0.435      \\
        & $ Dim = 256$         & 0.565 & 3.891 & 0.399 & \textbf{0.421}      \\
        & $ Dim = 768$         & 0.567 & 3.859 & 0.416 & 0.444      \\
        & Default      & \textbf{0.574} & \textbf{3.760} & \textbf{0.383} &0.424  \\
    \bottomrule
    \end{tabular}
    }
\end{table}

\subsection{Application to Epidemic Transmission}

The predicted mid-term trajectory can be applied to urban management applications. To demonstrate its practicality, we use the example of epidemic transmission analysis. 
Urban areas are typically divided into multiple subspaces. Researchers employ epidemic transmission models to simulate the infection dynamics within each subspace. Meanwhile, individual trajectories are utilized to guide the movement of urban residents, reproducing the propagation of infection between different subspaces. Comparing the discrepancies in epidemic transmission dynamics between observed and predicted trajectories provides insights into the model's predictive capacity for human mobility.

The SEIR model is commonly used to represent the epidemic spreading in a population~\cite{lai2020effect,chang2021mobility}. It categorizes all individuals into four states: susceptible (S), exposed (E), infectious (I), and removed (R). Susceptible individuals can become infected through close contact with those carrying the infection and subsequently transition into the exposed state. There exists a probability of transitioning from the exposed state to infection. Lastly, infectious individuals transition into the removed state with a probability. The removed state assumes that individuals are no longer infected and cannot transmit the infection to others.
Specifically, the probability of being exposed is formulated as $\rm P_{exp, i} = \frac{\alpha\beta I_i}{\rm N_i}$, where $\alpha$ is the number of close contacts, $\beta$ is the infectious rate, $\rm I_i$ and $\rm N_i$ stand for the number of infectious people and total population in the $i$-th subregion of the city. The infectious probability $\rm P_{inf}$ is the inverse of the average incubation period, while the removed probability $\rm P_{rem}$ is the inverse of the infectious period. The detailed parameter settings are illustrated in Table~\ref{tab:seir}. $\rm R_0$, the basic reproduction number, represents the average number of secondary infections caused by an infectious individual. It is calculated as the product of $\alpha$,$ \beta$, and the average infectious period. 

\begin{table}[ht]
\caption{Parameter settings in SEIR model.}
\label{tab:seir}
    \centering
    \resizebox{0.8\linewidth}{!}{
        \begin{tabular}{ll|ll}
        \toprule
        Parameter                       & Value         & Parameter             & Value                  \\
        \midrule
        $\rm \alpha$                    & 0.4           & $\rm \beta$           & 0.1                    \\
        Incubation Period               & 3*24        & $\rm P_{inf}$         & $\frac{1}{3\times 24}$ \\
        Infection Period                & 7*24        & $\rm P_{rem}$         & $\frac{1}{7\times 24}$ \\
        $\rm R_0$                           & 6.72      &                       &                        \\
        \bottomrule
        \end{tabular}
    }
\end{table}

In this study, due to the sparse population in the dataset, we initially aggregate grid-level trajectories to the admin level (census tract for \textit{Boston})  and assign an SEIR epidemic model to each administrative area. Subsequently, we employ these trajectories to construct a population transition probability matrix for each time window. These matrices are normalized to represent the probability of individuals choosing to stay or move between administrative areas during different periods. The population of each administrative area is obtained from Worldpop~\footnote{https://www.worldpop.org/}. For each time window, we first employ the SEIR model to calculate the status change of individuals within each administrative area, and then determine individual movement based on the transition probability matrix. All individuals are considered independent of each other. 

\begin{figure}[h]
    \centering
    \includegraphics[width=0.48\textwidth]{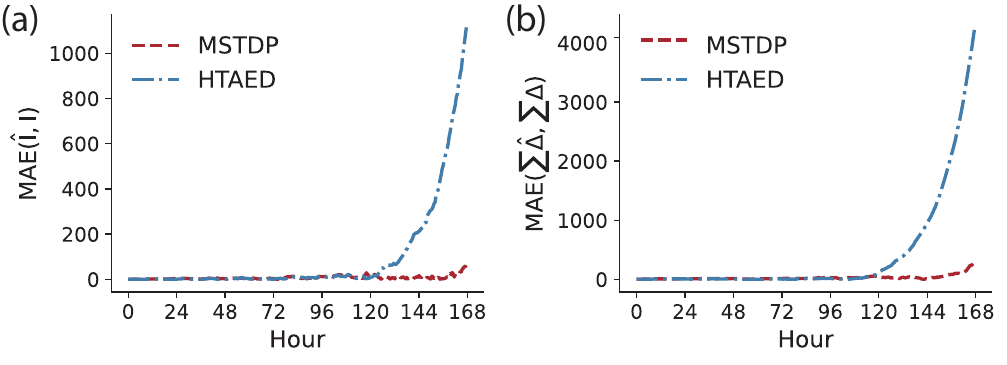}
    \caption{MAE over time. (a) The current active cases ($\rm I$); (b) The cumulative new cases ($\rm \sum\Delta$).}
    \label{fig:infection}
\end{figure}

We conduct 100 independent experiments using the actual data on \textit{Boston} dataset and the trajectories from seven-day consecutive forecasting with models. In each experiment, we randomly initialize 1000 infected individuals and monitor the half-hour fluctuations in the current active cases ($\rm I$) and cumulative new cases ($\rm \sum\Delta$) throughout the week. These predictive models are evaluated by comparing the infection numbers derived from the real and the predicted trajectories. The top-performing baseline, HTAED, is selected for comparison. We apply Mean Absolute Error (MAE) as the evaluation metric, and the average MAE is shown in Figure~\ref{fig:infection}. 

The experimental results indicate that the MAE of all models is comparatively low in the initial three days due to the setting of a three-day incubation period. However, beyond the fourth day, the MAE curve of the baseline model exhibits increasing divergence, while the MSTDP model consistently maintains a lower MAE. This difference can be attributed to the enhanced predictive capabilities of the MSTDP model in comparison to the baseline model in forecasting travel patterns. 
We calculate the half-hour average MAE for the seventh day for numerical comparison. The MAE ($\rm I$) values for MSTDP and HTAED are 17.1 and 138.5, while the MAE ($\rm \sum\Delta$) values are 96.2 and 258.8. The performance improvement of MSTDP is 87.6\% and 62.8\%, respectively. HTAED incurs a larger MAE due to its prediction of increased travel behaviors, which accelerates virus transmission and consequently leads to higher infection rates.

\subsection{Training and Inference Time Comparison}

\begin{figure}[h]
    \centering
    \includegraphics[width=0.48\textwidth]{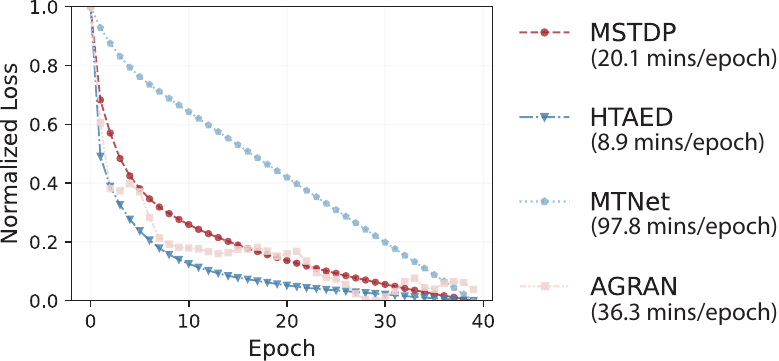}
    \caption{Comparison of loss reduction across epochs on \textit{Boston} dataset. The loss values are min-max normalized for better visualization.}
    \label{fig:loss}
\end{figure}

To evaluate the practical efficiency of these models, we compared the training and inference times on the Boston dataset. The experiments were conducted on an NVIDIA 4090 GPU, with the model configurations kept consistent with those used in the performance comparison experiments. Figure~\ref{fig:loss} presents the training time results, while Table~\ref{tab:infer_time} summarizes the inference time comparison. For inference time, we measured the duration required to predict daily mobility for 1,000 users with a batch size of 50.
Results demonstrate that HTAED achieves the fastest training and inference times, owing to its lightweight model structure. MSTDP significantly reduces training time by 79.4\% and 44.6\% compared to MTNet and AGRAN, respectively. Similarly, MSTDP achieves impressive inference time reductions of 99.5\% and 99.8\% compared to MTNet and AGRAN. These findings highlight MSTDP’s ability to balance computational efficiency with high predictive performance.

\begin{table}[ht]
\caption{Comparison of inference time on \textit{Boston} dataset.}
\label{tab:infer_time}
    \centering
    \resizebox{\linewidth}{!}{
    \begin{tabular}{c|cccc}
    \toprule
          & MSTDP & HTEAD & MTNet & AGRAN \\
    \midrule
        Infer time (s) & 10.1 & 1.4 & 1956.0 & 6853.7 \\
    \bottomrule
    \end{tabular}
    }
    
\end{table}

\section{Conclusion}
This study addresses mid-term human mobility prediction, a crucial task for understanding the daily travel patterns among urban residents and advancing urban science applications. We propose a novel approach, Multi-Scale Spatial-Temporal Decoupled Predictor (MSTDP), which employs a spatial-temporal decoupler and a hierarchical encoder to capture transition patterns across multiple time scales. Furthermore, we introduce a spatial heterogeneous graph to effectively represent the flow and spatial relationships between locations, facilitating the learning of semantic-rich location representation. We conduct extensive experiments to analyze the model's performance and demonstrate its effectiveness. Specifically, we evaluate MSTDP against nine baselines using five urban-scale mobile record datasets, demonstrating its superior performance. The statistical analysis, using Boston as a case study, examines key factors such as travel motifs, distances, and origin-destination flows in predicted trajectories. The results highlight that existing baseline models fail to accurately replicate the macro-level statistical patterns of individual mobility, whereas our proposed model effectively captures these patterns, underscoring its superior capability in modeling urban travel dynamics. Additionally, we apply the predictions to simulate epidemic dynamics with the SEIR model, using population transfer matrices derived from travel trajectories. The results show substantial improvements over the baselines, with reductions in MAE of 87.6\% for cumulative new cases and 62.8\% for active cases, highlighting MSTDP’s superior predictive capabilities in mobility modeling.

While our study has made significant progress in addressing mid-term mobility prediction, there remain several important directions for future exploration. First, MSTDP achieves strong performance on regular movement patterns and dense datasets; however, its generalization to irregular movement scenarios or sparse data environments presents opportunities for further improvement. Investigating adaptive modeling strategies, such as dynamic learning mechanisms or the incorporation of additional contextual factors (e.g., environmental or social information), could enhance the versatility and robustness of the model in such conditions. Second, the use of mobile phone data in trajectory prediction inherently involves sensitive user information, raising potential privacy concerns. Although all datasets in this study were anonymized, future work could explore integrating advanced privacy-preserving mechanisms to ensure robust data security during deployment in real-world applications. Third, real-world mobility behaviors are shaped by various external influences, such as weather patterns, public health emergencies, holidays, and major events. Future efforts could integrate these external data sources and develop dynamic mechanisms to adapt predictions based on such factors, thereby improving the model's applicability in diverse and evolving scenarios. Finally, an important aspect of mid-term mobility prediction is the challenge of cumulative error propagation in multi-step predictions, which impacts accuracy over longer time horizons. Future research could focus on designing methodologies to further enhance the accuracy and reliability of multi-day trajectory forecasting.

\bibliographystyle{IEEEtran}
\bibliography{reference}

\end{document}